\pdfoutput=1
\documentclass[11pt,a4paper]{article}
\usepackage[hyperref]{emnlp2020}
\usepackage{times}
\usepackage{latexsym}

\usepackage{microtype}

\usepackage{linguex}
\usepackage[export]{adjustbox}
\usepackage{textcomp}

\aclfinalcopy 
\def\aclpaperid{2512} 

\title{QADiscourse - Discourse Relations as QA Pairs:
Representation, Crowdsourcing and Baselines}

\author{First Author \\
  Affiliation / Address line 1 \\
  Affiliation / Address line 2 \\
  Affiliation / Address line 3 \\
  \texttt{email@domain} \\\And
  Second Author \\
  Affiliation / Address line 1 \\
  Affiliation / Address line 2 \\
  Affiliation / Address line 3 \\
  \texttt{email@domain} \\} 
  
\author{Valentina Pyatkin\textsuperscript{1} \, Ayal Klein\textsuperscript{1} \, Reut Tsarfaty\textsuperscript{1,2} \, Ido Dagan\textsuperscript{1}\\
\textsuperscript{1}Computer Science Department, Bar Ilan University \\
\textsuperscript{2}Allen Institute for Artificial Intelligence \\
  {\tt  \{valpyatkin,ayal.s.klein,ido.k.dagan,reut.tsarfaty\}@gmail.com}
  }

\date{}

\begin{document}
\maketitle
\begin{abstract}
Discourse relations describe how two propositions relate to one another, and identifying them automatically is an integral part of natural language understanding.  However, annotating discourse relations typically requires expert annotators. Recently, different semantic aspects of a sentence have been represented and crowd-sourced via question-and-answer (QA) pairs. This paper proposes a novel representation of discourse relations as QA pairs,  which in turn allows us to crowd-source wide-coverage data annotated with discourse relations,  via an intuitively appealing interface for composing such questions and answers. Based on our proposed  representation, we collect a novel and wide-coverage QADiscourse dataset, and present baseline algorithms for predicting QADiscourse relations.
\end{abstract}

\section{Introduction}
Relations between propositions are commonly termed {\em Discourse Relations}, and their importance to the automatic understanding of the content and structure of narratives has been extensively studied  \cite{grosz1986attention, asher2003logics, webber2012discourse}.
The automatic   parsing of such relations  is thus relevant to multiple areas of NLP research,  from extractive tasks such as document summarization  to automatic analyses of narratives and event chains \cite{li2016role,lee2019multi}.

\begin{table}
\resizebox{\columnwidth}{!}{%
    \centering
    \small
    \begin{tabular}{|p{2.85in}|} \hline
       The executions were \textbf{spurred} by lawmakers \textbf{requesting} action to \textbf{curb} rising crime rates.\\
       \hline
        \textcolor{blue}{\emph{What is the reason} lawmakers \textbf{requested} action?} \textcolor{brown}{to \textbf{curb} rising crime rates} \\
        \textcolor{blue}{\emph{What is the result} of lawmakers \textbf{requesting} action to \textbf{curb} rising crime rates?} \textcolor{brown}{the executions were \textbf{spurred}} \\ \hline
               I \textbf{decided} to do a press conference [...], and I \textbf{did} that going into it \textbf{knowing} there would be consequences.\\
       \hline
        \textcolor{blue}{\emph{Despite what} did I \textbf{decide} to do a press conference?} \textcolor{brown}{\textbf{knowing} there would be consequences} \\ \hline
    \end{tabular}}
        \caption{Sentences with their corresponding Question-and-Answer pairs. The bottom example shows how {\em implicit} relations are captured as  QAs.}
    \label{fig1}
\end{table}

So far, discourse annotation has been mainly conducted by experts, relying on carefully crafted linguistic schemes. Two cases in point  are PDTB \cite{prasad2008penn, webber2019penn} and RST \cite{mann1988rhetorical, carlson2002rst}. Such annotation however is slow and costly. Crowd-sourcing discourse relations, instead of using experts, can be very useful for obtaining  larger-scale  training data for discourse parsers. 

One plausible way for acquiring linguistically meaningful annotations from laymen is using the relatively recent QA methodology, that is, converting a set of linguistic concepts to intuitively simple Question-and-Answer (QA) pairs. Indeed, casting the semantic annotations of individual propositions as narrating a QA pair gained increasing attention in recent years, ranging from QA driven Semantic Role Labeling (QASRL) \cite{he2015question, fitzgerald2018large, roit2019crowdsourcing} to covering all semantic relations as in QAMR \cite{michael2018crowdsourcing}.  These representations were also shown to improve downstream tasks, for example by providing indirect supervision for recent MLMs \cite{he2019incidental}. 

In this work we address the challenge of crowd-sourcing information of  higher complexity,  that of discourse relations, using QA pairs. We present the QADiscourse approach to representing intra-sentential Discourse Relations as QA pairs, and we show that with an appropriate crowd-sourcing setup, complex relations between clauses can be effectively recognized by non-experts. This layman annotation could also easily be ported to other languages and domains.

Specifically, we define the QADiscourse task to be the detection of the two discourse units, and the labeling of the relation sense between them.
The two units are represented in the question body and the answer, respectively, while the question type, as expressed by its prefix, represents the discourse relation sense between them. This representation is illustrated in Table~\ref{fig1} and the types of questions we focus on are detailed in Table~\ref{mapping}. This scheme 
allows us to ask about 
both {\em explicit} and {\em implicit} relations. To our knowledge, there has been no work on collecting such question types in a systematic way.

The contribution of this paper is thus manifold. (i) We propose a novel QA-based representation for  discourse relations reflecting a subset of the sense taxonomy of PDTB 3.0 \cite{webber2019penn}.
(ii) We propose an 
annotation methodology to crowd-source such discourse-relations QA pairs. And, (iii) given this representation and  annotation setup, we collected QADiscourse annotations for about 9000 sentences, resulting in more than 16600 QA pairs, which we will openly release. 
Finally, (iv) we implement a QADiscourse parser, serving as a baseline for predicting discourse questions and respective answers, capturing multiple discourse-based propositions in a sentence.

\section{Background}

\paragraph{Discourse Relations}
Discourse Relations in the Penn Discourse Treebank (PDTB) \cite{prasad2008penn, webber2019penn}, as seen in ex. \ref{PDTBex}, are represented by two arguments, labeled either \textit{Arg1} or \textit{Arg2}, the discourse connective (in case of an explicit relation) and finally the relation sense(s) between the two, in this case both Temporal.Asynchronous.Succession and Contingency.Cause.Reason.
\ex. \label{PDTBex} \underline{BankAmerica climbed 1 3/4 to 30} (\textit{Arg1}) \textbf{after} \textit{PaineWebber boosted its investment opinion on the stock to its highest rating} (\textit{Arg2}).

These relations are called {\em shallow} discourse relations since, contrary to the Rhetorical Structure Theory (RST) \cite{mann1988rhetorical, carlson2002rst}, they do not recursively build a tree.
The PDTB organizes their sense taxonomy, of which examples can be seen in Table~\ref{mapping}, into three levels, with the last one denoting the direction of the relation.
The PDTB annotation scheme has additionally been shown to be portable to other languages \cite{Zeyrek2018, long2020shallow}.

\paragraph{Semantic QA Approaches}
\label{qaapproaches}
Using QA structures to represent semantic propositions has been proposed as a way to generate ``soft" annotations, where the resulting representation is formulated using natural language, which is shown to be more intuitive for untrained annotators \cite{he2015question}. This allows much quicker, more large-scale annotation processes \cite{fitzgerald2018large} and when used in a more controlled crowd-sourcing setup, can produce high-coverage quality annotations \cite{roit2019crowdsourcing}. 

As displayed in Table~\ref{qamr_and_qasrl}, both QASRL and QAMR collect a set of QA pairs, each representing a single proposition, for a sentence. In QASRL the main target is a predicate, which is emphasized by replacing all content words in the question besides the predicate with a placeholder. The answer constitutes a span of the sentence. The annotation process itself for QASRL is very controlled, by suggesting questions created with a finite-state automaton. QAMR, on the other hand, allows to freely ask all kinds of questions about all types of content words in a sentence.

\begin{table}
\resizebox{\columnwidth}{!}{%
    \centering
    \small
    \begin{tabular}{|p{2.85in}|} \hline
       \underline{QASRL}: Back in Warsaw that year, Chopin \textbf{heard} Niccolò Paganini \textbf{play} the violin, and \textbf{composed} a set of variations, Souvenir de Paganini.\\
       \hline
        \textcolor{blue}{What did someone \textbf{hear}?} \textcolor{brown}{Niccolò Paganini play the violin}\\
        \textcolor{blue}{When did someone \textbf{compose} something?} \textcolor{brown}{that year}\\\hline
       \underline{QAMR}: Actor and television host Gary Collins died yesterday at age 74.\\
       \hline
        \textcolor{blue}{What kind of host was Collins?} \textcolor{brown}{television} \\ 
                \textcolor{blue}{How old was Gary Collins?} \textcolor{brown}{74} \\ 
        \hline
    \end{tabular}}
    \caption{Examples of QASRL and QAMR annotations.}
    \label{qamr_and_qasrl}
\end{table}

\paragraph{QA Approaches for Discourse}
The relation between discourse structures and questioning has been pointed out by \citet{van1995discourse}, who claims that the discourse is driven by explicit and implicit questions:
a writer carries a topic forward by answering anticipated questions given the preceding context. \citet{roberts2012information} introduces the term \textit{Question Under Discussion} (QUD), which stands for a question that interlocuters accept in discourse and engage in finding its answer. More recently, there has been work on annotating QUDs, by asking workers to identify questions raised by the text and checking whether or not these raised questions get answered in the following discourse \cite{westera2019asking, westera2020ted}. These QUD annotations are conceptually related to QADiscourse by representing discourse information through QAs, solicited from laymen speakers. The main difference lies in the propositions captured: we collect questions that have an answer in the sentence, targeting specific relation types. In the QUD annotations \cite{westera2020ted} any type of question can be asked that might or might not be answered in the following discourse.

\paragraph{Previous Discourse Parsing Efforts}
\label{prevparsers}
Most of the recent work on models for (shallow) discourse parsing focuses on specific subtasks, for example on argument identification \cite{knaebel2019window}, or  discourse sense classification \cite{dai2019regularization, shi2019next, van2019employing}.
Full (shallow) discourse parsers tend to use a pipeline approach, for example by having separate classifiers for implicit and explicit relations \cite{lin2014pdtb}, or by building different models for intra- vs. inter-sentence relations \cite{biran2015pdtb}. We also adopt the pipeline approach for our baseline model (Section \ref{baselinemodel}), which performs both relation classification and argument identification, since our QA pairs jointly represent arguments and relations.

\paragraph{Previous Discourse Crowdsourcing Efforts}
There has been research on how to crowd-source discourse relation annotations. \citet{kawahara2014rapid} crowd-source Japanese discourse relations and simplify the task by reducing the tagset and extracting the argument spans automatically. A follow-up paper found that the data quality of these Japanese annotations was lacking compared to expert annotations \cite{kishimoto2018improving}. Furthermore, \citet{yung2019crowdsourcing} even posit that it is impossible to crowdsource high quality discourse sense annotations and they suggest to re-formulate the task as a discourse connective insertion problem. This approach has previously also been used in other configurations \cite{rohde2016filling, scholman2017crowdsourcing}. Similarly to our QADiscourse approach, inserting connectives also works with soft natural language annotations, as we propose, but it simplifies the task greatly, by only annotating the connective, rather than retrieving complete discourse relations.

\section{Representing Discourse Relations as QA pairs}
In this section we discuss how to represent shallow {\em discourse relations} through
QA pairs. For an overview, consider the second sentence in Table \ref{fig1}, and the two predicates `decided' and `knowing', each being part of a discourse unit. The sense of the discourse relation between these two units can be characterized by the question prefix ``Despite what ...?" (see Table \ref{mapping}). Accordingly, the full QA pair represents the proposition asserted by this discourse relation, with the question and answer corresponding to the two discourse units. A complete QADiscourse representation for a text would thus consist of a set of such QAs, representing all propositions asserted through discourse relations.

\paragraph{The Discourse Relation Sense}
We want our questions to denote relation senses.
To define the set of discourse relations covered by our approach, we derived a set of question templates that cover most discourse relations in the PDTB 3.0 \cite{webber2019penn, prasad2008penn}, as shown in Table~\ref{mapping}. 
Each question template starts with a question prefix, which specifies the relation sense. The placeholder X is completed to capture the discourse unit referred to by the question, as in Table~\ref{fig1}.

Few PDTB senses are not covered by our question prefixes. First, senses with pragmatic specifications like \textit{Belief} and \textit{SpeechAct} were collapsed into their general sense. Second, three \textit{Expansion} senses were not included because they usually do not assert a new ``informational" proposition, about which a question could be asked, but rather capture structural properties of the text. One of those is \textit{Expansion.Conjunction}, which is one of the most frequently occurring senses in the PDTB, especially in intra-sentential VP conjunctions, where it makes up about 70\% of the sense instances \cite{webber2016discourse}. Ex. \ref{excluded_relations} displays a discourse relation with two senses, one of which \textit{Expansion.Conjunction}. While it is natural to come up with a question targeting the causal sense, 
the conjunction relation does not seem to assert any proposition about which an informational question may be asked. 
\ex.\label{excluded_relations}``Digital Equipment announced its first mainframe computers, \emph{targeting IBM's \textbf{largest} market} \underline{and} \emph{heating up the industry's \textbf{biggest} rivalry}." (explicit: \textit{Expansion.Conjunction}, implicit: \textit{Contingency.Cause.Result})

Finally, we removed \textit{Purpose} as a (somewhat subtle) individual sense and subsumed it with our two causal questions.

\begin{table}
\resizebox{\columnwidth}{!}{%
    \centering
    \small
\begin{tabular} {l r } \hline
 PDTB Sense & Question Template \\
\hline
Expansion.Substitution & Instead of what X?\\ 
Expansion.Disjunction &  What is an alternative to X?\\
Expansion.Exception &  Except when X?\\
Comparison.Concession &  Despite what X?\\
Comparison.Contrast &  What is contrasted with X?\\
Expansion.Level-of-detail &  What is an example of X?\\
Comparison.Similarity &  What is similar to X?\\
Temporal.Asynchronous &  After/Before what X?\\
 &  Until/Since when X?\\
Temporal.Synchronous &  While what X?\\
Contingency.Condition & In what case X?\\
Contingency.Neg.-cond. &  Unless what X?\\
Expansion.Manner &  In what manner X?\\
Contingency.Cause &  What is the result of X?\\
 &  What is the reason X \\
 \hline
\end{tabular}}
\caption{Informational PDTB senses mapped to our question templates.}
\label{mapping}
\end{table}

Our desiderata for the question templates are as follows. 
First, we want the question prefixes to be applicable  to as many scenarios as possible in which discourse relations can occur, while at the same time ideally adhering to a one-to-one mapping of sense to question. 
Similarly, we avoid question templates that are too general. The \textsc{When}-Question in QASRL (Table~\ref{qamr_and_qasrl}), for instance, can be used for either Temporal or Conditional relations. Here we employ more specific question prefixes to remove this ambiguity. Finally, as multiple relation senses can hold between two discourse units \cite{rohde2018discourse}, we similarly allow multiple QA pairs for the same two discourse units. 

\paragraph{The Discourse Units}
The two sentence fragments, typically clauses, that we relate with a question are the discourse units. In determining what makes a discourse unit, we include verbal predicates, noun phrases and adverbial phrases as potential targets. This, for example, would also cover such instances: ``Because of \underline{the rain} ..." or ``..., albeit \underline{warily}". We call the corresponding verb, noun or adverb heading a discourse unit a \textit{target}.

A question is created by choosing a question prefix, an auxiliary, if necessary, and copying words from the sentence. It can then be manually adjusted to be made grammatical. Similarly, the discourse unit making up the answer consists of words copied from the sentence and can be modified to be made grammatical. 
Our question and answer format thus deviates considerably from the  QASRL representation. By not introducing placeholders, questions sound more natural and easy to answer compared to QASRL, while still being more controlled than the completely free form questions of QAMR. In addition, allowing for small grammatical adjustments introduces valuable flexibility which contribute to the intuitiveness of the annotations.

\paragraph{Relation Directionality}
Discourse relations are often directional. 
Our QA format introduces directionality by placing discourse units into either the question or answer. In some question prefixes, a single order is dictated by the question. As seen in ex.\ 1 of Table~\ref{figsymmetric}, because the question asks for the \textit{condition}, the condition itself will always be in the answer.
Another ordering pattern occurs for \textit{symmetric} relations, meaning that the relation's assertion remains the same no matter how the arguments are placed into the question and answer, as in ex.\ 2 in Table~\ref{figsymmetric}.
Finally, certain pairs of relation senses are considered \textit{reversed}, such as for causal (reason vs. result) and some of the temporal (before vs. after) question prefixes. In this case, two QA pairs with different question prefixes can denote the same assertion when the target discourse units are reversed, as shown in ex.\ 3 in Table~\ref{figsymmetric}.
These patterns of directionality impact annotation and evaluation, as would be described later on.

\begin{table}\resizebox{\columnwidth}{!}{%
    \centering
    \small
            \begin{tabular}{|p{2.85in}|} \hline
1. And I also \textbf{feel} like in a capitalistic \textbf{society}, checks and balances \textbf{happen} when there \textbf{is} competition. \\
       \hline
        \textcolor{blue}{\textbf{In what case} do checks and balances happen?} \textcolor{brown}{when there is competition in a capitalistic society} \\\hline
2. Whilst \textbf{staying} in the hotel, the Wikimedian group \textbf{met} two MEPs who \textbf{chose} it in-preference to dramatically more-expensive Strasbourg accommodation. \\
       \hline
        \textcolor{blue}{\textbf{What is contrasted with} the hotel?} \textcolor{brown}{dramatically more-expensive Strasbourg accommodation} \\ \hline
        3. There were no fare hikes \textbf{announced} as both passenger and freight fares had been \textbf{increased} last month. \\
       \hline
        \textcolor{blue}{\textbf{What is the reason} there were no fare hikes announced?} \textcolor{brown}{as both passenger and freight fares had been increased last month} \\ \hline
                \textcolor{blue}{\textbf{What is the result of} both passenger and freight fares having been increased last month?} \textcolor{brown}{There were no fare hikes announced} \\ \hline
    \end{tabular}}
    \caption{Example sentences with their corresponding Question-and-Answer pairs.}
    \label{figsymmetric}
\end{table}

\section{The Crowdsourcing Process}
\label{Process}

\paragraph{Pool of Annotators}
To find a suitable group of annotators we followed the Controlled Crowdsourcing Methodology of \citet{roit2019crowdsourcing}. We first released two trial tasks, after which we selected the best performing workers. These  workers then underwent two short training cycles, estimated to take about an hour each, which involved reading the task guidelines, consisting of 42 slides\footnote{\url{https://github.com/ValentinaPy/QADiscourse}}, completing 30 HITs per round and reading personal feedback after each round (preparing these feedbacks consumed about 4 author work days). 11 workers successfully completed the training. 

For collecting production annotations of the Dev and Test Sets, each sentence was annotated by 2 workers independently, followed by a third worker who adjudicated their QA pairs to produce the final set. For the Train Set, sentences were annotated by a single worker, without adjudication.

\paragraph{Preprocessing}
\label{targetextraction}
In preprocessing, question targets are extracted automatically using heuristics and POS tags: a sentence is segmented using punctuation and discourse connectives (from a list of connectives from the PDTB). For each segment, we treat the last verb in a consecutive span of verbs as a separate target. In case a segment does not contain a verb, but does start with a discourse connective, we choose one of the nouns (or adverbs) as target. The following illustrates our target extraction: 
\label{ex1}\textit{[Despite labor-shortage \textbf{warnings},] [80\% \textbf{aim} for first-year wage increases of under 4\%;] [and 77\% \textbf{say} they'd try to \textbf{replace} workers,] [if \textbf{struck},] [or would \textbf{consider} it.]}

\paragraph{Annotation Tool and Procedure}
\label{anno}
Using Amazon Mechanical Turk, we implemented two interfaces\footnote{Please refer to the appendix for all UI screenshots.}, one for the \textit{QA generation} and one for the \textit{QA adjudication} step. 

In the {\em QA generation} interface, workers are shown a sentence with all target words in bold. 
Workers are instructed to generate one or more questions that relate two of these target words. The question is generated by first choosing a question prefix, then, if applicable, an auxiliary, then selecting one or more spans from the sentence to form the complete question, and lastly, change it to make it grammatical. Given the generated question, the next step involves answering that question by selecting span(s) from the sentence. Again, the answer can also be amended to be made grammatical. 

The {\em QA adjudication} interface displays a sentence and all the QA pairs produced for that sentence by two annotators. For each QA pair the adjudicator is asked to either label it as \textit{correct}, \textit{not correct} or \textit{correct, but not grammatical}. Duplicates and nonsensical QA pairs labeled as \textit{not correct} are omitted from the final dataset. As a last step, the first author manually corrected all the \textit{not grammatical} instances. 

\paragraph{Data and Cost}
We sampled sentences from the Wikinews and Wikipedia sections of Large Scale QASRL \cite{fitzgerald2018large} while following their Train, Dev \& Test splits.
Descriptive statistics for the final dataset are shown in Table~\ref{datastats} and \ref{datacounts}.

Annotating a sentence of Dev and Test yielded 2.11 QA pairs with a cost of 50.3\textcent \space on average.
For Train, the average cost was 37.1\textcent \space for 1.72 QAs per sentence.

\begin{table}[t]
\centering
\small
\begin{tabular}{lcc} \hline
 Dataset Split & Sentences & Questions  \\ \hline
Wikinews Train & 3098 & 4760 \\ 
Wikinews Dev & 669 &  1108 \\ 
Wikinews Test & 658 & 1498 \\
Wikipedia Train & 3277 & 6225 \\ 
Wikipedia Dev & 667 & 1524 \\ 
Wikipedia Test & 678 & 1498 \\ 
\hline
Overall & 9047 & 16613 \\
\hline
\end{tabular}
\caption{Dataset Statistics: Number of sentences containing at least 1 QA annotation and the total number of QAs collected.}
\label{datastats}
\end{table}

\begin{table}[tb!]
\centering
\small
\scalebox{0.9}{
\begin{tabular}{lll} \hline
 QA Prefix & Count & Proportion \\ \hline
In what manner X? &  4225 & 25\%\\ 
What is the reason X? & 3238 & 19\%\\
What is the result of X? & 2735 & 16 \%\\
What is an example of X? & 1757& 11 \%\\
After what X? & 1099& 7 \%\\
While what X? & 1060& 6 \%\\
In what case X? &  509& 3 \%\\
Despite what X? & 477& 3 \%\\
What is contrasted with X? & 317& 2 \%\\
Before what X? & 299& 2 \%\\
Since when X? & 279& 2 \%\\
What is similar to X? & 218& 1 \%\\
Until when X? &  155& 1 \%\\
Instead of what X? & 105& 1 \%\\
What is an alternative to X? &  92 & $\leq$ 1 \%\\
Except when X? &  27 &  $\leq$ 1 \%\\
Unless what X? &  21 &  $\leq$ 1\%\\
\hline
\end{tabular}}
\caption{Counts of collected question types.}
\label{datacounts}
\end{table}

\section{Dataset Evaluation}

\subsection{Evaluation Metrics}
\label{evalmetrics}
We aim to evaluate QA pairs, as the output of both the annotation process and the question generation and answering model, which are not the same as discourse relation triplets. There are multiple difficulties that arise when evaluating the QADiscourse setup. We allow multiple labels per proposition pair and thus need evaluation measures suitable for multi-label classification. Annotators are generating the questions and answers, which contrary to a pure categorical labelling task implies that we have to take into consideration question and answer paraphrasing and natural language generation inconsistencies. This requires us to use metrics that create alignments between sets of QAs, which means that existing discourse relation evaluation methods, such as from CoNLL-2015 \cite{xue2015conll}, are not applicable. The following metrics, which we apply for both the quality analysis of the dataset and the parser evaluation, are closely inspired by previous work on collecting semantic annotations with QA pairs \cite{roit2019crowdsourcing, fitzgerald2018large}. 

\paragraph{Unlabeled Question and Answer Span Detection (UQA) (F1)}
This metric is inspired by the question alignment metric for QASRL, which takes into account that there are many ways to phrase a question and therefore an exact match metric will be too harsh. Given a sentence and two sets of QA pairs produced for that sentence, such as gold and predicted sets, we want to match the QAs from the two sets for comparison. A QA pair is aligned with another QA pair that has the maximal \emph{intersection over union} (IOU) $\geq$ 0.5 on a token-level, or else remains unaligned\footnote{The average length for tokenized questions and answers is 12.22 and 10.27 respectively.}. Since we allow multiple QA pairs for two targets, we also allow one-to-many and many-to-many alignments. As we are evaluating unlabeled relations at this point, we do not consider relation direction and therefore do not differentiate between question and answer spans.

\paragraph{Labeled Question and Answer Span Detection (LQA) (Accuracy)}
Given the previously produced alignments from UQA we check for the exact match of aligned question prefixes. For many-to-many and many-to-one alignments we count as correct if there is overlap of at least one question prefix. \textit{Reversed} and \textit{symmetric} question prefixes are converted to a more general label for fair comparison. 

\subsection{Dataset Quality}
\paragraph{Inter-Annotator Agreement (IAA)}
To calculate the agreement between individual annotators we use the above metrics (UQA and LQA) for different worker-vs-worker configurations. The setup is the following: A set of 4 workers annotates the same sentences (around 60), from which we then calculate the agreement between all the possible pairs of workers. We repeat this process 3 times and show the average agreement scores in Table~\ref{IAA}.
The scores after adjudication, pertaining to the actual dataset agreement level, are produced by comparing the resulting annotation of two worker triplets, each consisting of two annotators and a separate adjudicator on the same data, averaged over 3 samples of 60 sentences each. These results show that adjudication notably improves agreement.
\begin{table}[tb!]
\centering
\small
\begin{tabular}{ccc} \hline
 & UQA & LQA  \\ \hline
Before Adjud. & 76.87 &  56.64  \\ 
After Adjud. & 87.44 & 65.46 \\ \hline
\end{tabular}
\caption{IAA scores before and after adjudication.}
\label{IAA}
\end{table}

\subsection{Agreement with Expert Set}
Our Expert set consists of 25 sentences annotated with QA pairs by the first author of the paper. Comparing the adjudicated crowdsourced annotations with the Expert Set yields a UQA (LQA) of \textbf{93.9} (\textbf{80}), indicating a high quality of our collected annotations.
The main issue in disagreement arises from sentences that do not contain overt propositional discourse relations, where workers attempt to ask questions anyways, resulting in sometimes unnatural or overly implicit questions.

\subsection{Comparison with PDTB}
We crowdsourced QA annotations of 60 sentences from section 20 of the PDTB (commonly used as Train) with our QA annotation protocol.
The PDTB arguments are aligned with the QA-pairs using the UQA metric, by considering \textit{Arg1} and \textit{Arg2} as the text spans to be aligned with the question and answer text\footnote{Such alignment is usually straightforward, since annotators do not add content words when producing QAs.}, yielding \textbf{83.2} Precision, \textbf{87.5} Recall and an F1 of \textbf{85.3}.

A manual comparison of the PDTB labels with the Question Prefixes reveals that in most of the cases the senses overlap in meaning, with some exceptions on both sides. 60\% of aligned annotations correspond exactly in the discourse relation sense they express. The remaining 40\% of the QA-pairs belong to either of the following categories: 

(1) Discourse relations that we deemed to be non-informational at the propositional level were many times still annotated with our QA pairs. Take this sentence: [...], \textit{a Soviet-controlled regime remains in Kabul}, the refugees sit in their camps, and \textit{the restoration of Afghan freedom seems as far off as ever}.
The PDTB posits an \textit{Exp.Conjunction} relation between the two cursive arguments, which is a relation type that we do not cover in the QA framework, yet our annotators saw an implied causal relation which they expressed with the following (sensible) QA pair:
\textbf{What is the reason} \textit{the restoration of Afghan freedom seems as far off as ever? - a soviet-controlled regime remains in Kabul}.

(2) Interestingly, we observe that some annotation decision difficulties described in the PDTB \cite{webber2019penn} are also mirrored in our collected data. One of those arising ambiguities is the difference between \textit{Comparison.Contrast} and \textit{Comparison.Concession}, in our case \textit{Despite what} and \textit{What is contrasted with}. In the manually analyzed data sample, 3 such confusions were found between the QADiscourse and the PDTB annotations.

(3) There were 15 instances of PDTB relation senses that were erroneously not annotated with an appropriate QA pair, even though a suitable Question Prefix exists, corresponding to some of the 12.5\% recall misses in the comparison. 

(4) On the contrary, there were 36 QA instances that capture appropriate propositions which were completely missed in the PDTB \footnote{The full list of these instances can be found in the appendix.}. For example, in Table~\ref{exmissed}, the PDTB only mentions the causal relation, while QADiscourse found both the causal and the temporal sense:
\begin{table}
\resizebox{\columnwidth}{!}{%
    \centering
    \small
    \begin{tabular}{|p{2.85in}|} \hline
Frank Carlucci III was named to this telecommunications company\'s board, filling the vacancy created by the death of William Sobey last May. (Contingency.Cause.Result)\\
       \hline
        \textcolor{blue}{\emph{After what} was Frank Carlucci III named to this telecommunications company's board?} \textcolor{brown}{the death of William Sobey last may} \\
        \textcolor{blue}{\emph{What is the reason} Frank Carlucci III was named to this telecommunications company's board?} \textcolor{brown}{filling the vacancy created by the death of William Sobey last May} \\ \hline
    \end{tabular}}
    \caption{Example of a QADiscourse relation which was not captured in the PDTB.}
    \label{exmissed}
\end{table}

Additionally we noticed that annotators tend to ask ``What is similar to..?” questions about conjunctions, indicating that conjoined clauses seem to imply a similarity between them, while the similarity relation in the PDTB is rather used in more explicit comparison contexts. The ``In what case..?" questions were sometimes used for adjuncts specifying a time or place. Overall, these comparisons show that agreement with the PDTB is good, with QADiscourse even finding additional valid relations, 
indicating that it is feasible to crowdsource high-quality discourse relations via QADiscourse.

\subsection{Comparison with QAMR and QASRL}
While commonly treated as two distinct levels of textual annotations, there are nevertheless some commonalities between shallow discourse relations and semantic roles. This interplay of discourse and semantics has also been noted by \citet{prasad2015bridging}, who made use of clausal adjunct annotations in PropBank to enrich intra-sentential discourse annotations and vice versa. Similarly, we found that there are questions in QASRL, QAMR and QADiscourse which express kindred relations: \textit{Manner}, \textit{Condition}, \textit{Causal} and \textit{Temporal} relations could all be asked about using QASRL-like \textsc{Wh-}Question. But then the point of question ambiguity arises: if ``When" can be used to ask about conditional relations, it is more often also used to denote temporal relations. This under-specification becomes problematic when attempting to map between QAs and labels from resources such as PropBank. Therefore, despite the propositional overlap of some of the question types, QADiscourse additionally enriches and refines QASRL annotations. 

Since QAMR does not restrict itself to predicate-argument relations only, we performed an analysis of whether annotators tend to ask about QADiscourse-type relations in a general QA setting.
965 sentences contain both QAMR and QADiscourse annotations, with 1505 QADiscourse pairs, of which we could align 101 (7\%) to QAMR annotations, using the UQA-alignment algorithm. We conclude that QAMR and QADiscourse target mostly different propositions and relation types.

\begin{table}[tb!]
\centering
    \small
\begin{tabular}{cc} \hline
 Question Prefix & Count \\ \hline
What is the reason/result of & 23/20   \\ 
In what manner & 19 \\
While/After/Before what & 19 \\
What is an example of & 10 \\
Since/Until when & 5 \\
In what case & 4 \\
Despite what & 1 \\
\hline
\end{tabular}
\caption{Count of QADiscourse Question Prefixes of questions that could be aligned to QAMR.}
\label{QAMR_Analysis}
\end{table}

Within the 101 QADiscourse QAs that were aligned with QAMR questions (Table~\ref{QAMR_Analysis}), 
causal and temporal relations are very common, usually expressed, as expected, by ``Why" or ``When" questions in QAMR. 
In other cases, the aligned questions express different relation senses. Notably, the QADiscourse \textit{In what manner} relation aligns with a ``How" QAMR question only once out of 19 cases. 
Often, it seems that QADiscourse annotators were tempted to ask a somewhat inappropriate \textit{manner} question while the relation between the predicate and the answer corresponded to a direct semantic role (like location) rather than to a discourse relation (first example in Table~\ref{qamr_aligned}).
The second example in Table~\ref{qamr_aligned} corresponds to a case where the predicate-answer relation has two senses, a discourse sense captured by QADiscourse 
(\textit{What is an example of}), as well as a semantic role (``theme"), captured by a ``What" question in QAMR. These observations suggest interesting future research on integrating QADiscourse annotations with semantic role QA annotations, like QASRL and QAMR.

\begin{table}
\resizebox{\columnwidth}{!}{%
    \centering
    \small
    \begin{tabular}{|p{2.85in}|} \hline
She said he ``had friends in every political party ..." \\
       \hline
\textsc{QADisc.}: \textcolor{blue}{\emph{In what manner} did he have friends?}\\
\textsc{QAMR}: \textcolor{blue}{\emph{Where} does he have friends?}\\ \textsc{Answer:} {\textcolor{brown}{in every political party}}\\\hline
... your internet access provider can still keep track of what websites you visit, websites can collect information about you and so on. \\
       \hline
\textsc{QADisc.}: \textcolor{blue}{\emph{What is an example of} something your internet access provider can still keep track of?} \\
\textsc{QAMR}: \textcolor{blue}{Your provider can keep track of \emph{what?}} \\ \textsc{Answer:} \textcolor{brown}{what websites you visit}\\ \hline
    \end{tabular}}
    \caption{Examples of interesting aligned cases between QAMR and QADiscourse.}
    \label{qamr_aligned}
\end{table}

\section{Baseline Model for QADiscourse}
\label{baselinemodel}

\begin{table*}[htb!]
\resizebox{\textwidth}{!}{%
    \centering
    \small
    \begin{tabular}{|p{3in}||p{3in}|} \hline
       \textbf{1.} This process, [...], rather than maintaining it as a network of unequal principalities, would ultimately be completed by Caesar's successor [...] & \textbf{4.} A writer since his teens, Pratchett first came to prominence with the Discworld novel [...]\\ \hline
       \textcolor{blue}{Instead of what would this process [...] be completed by Caesar's successor?} - \textcolor{brown}{rather than maintaining it as a network of unequal principalities} & \textcolor{blue}{Since when did Pratchett a writer?} - \textcolor{brown}{since his teens}\\\hline
               \textbf{2.} Most decked vessels were mechanized, but two-thirds of the open vessels were traditional craft propelled by sails and oars. &  \textbf{5.} Each segment of the search could last for several weeks before resupply in Western Australia.  \\ \hline
       \textcolor{blue}{What is contrasted with most decked vessels appearing mechanized?} - \textcolor{brown}{two-thirds of the open vessels were traditional craft propelled by sails and oars} &        \textcolor{blue}{What is the reason each segment of the search could last for several weeks?} - \textcolor{brown}{before resupply in Western Australia} \\\hline
       \textbf{3.} It could hit Hawaii if it stays on its predicted path.&        \textbf{6.} For Cook Island Maori , it was 29 \% compared to 23 \% ; for Tongans , 37 \% to 29 \% [...].\\ \hline
       \textcolor{blue}{In what case could it hit Hawaii?} - \textcolor{brown}{if it stays on its predicted path} &        \textcolor{blue}{What is contrasted with it For Cook Island Maori?} - \textcolor{brown}{23 \%} \\\hline
    \end{tabular}}
    \caption{Examples of the QA output of the full pipeline: On the left column successful predictions and on the right wrong predictions (4: not grammatical but sensible, 5: non-sensical but grammatical, 6: neither).}
    \label{genqex}
\end{table*}

In this section we aim to devise a baseline discourse parser based on our proposed representation, which accepts a sentence as input and outputs QA pairs for all discourse relations in that sentence, to be trained on our collected data. Similarly to previous work on discourse parsing (Section \ref{prevparsers}), our proposed parser is a pipeline consisting of three phases: {\em (i) question prefix prediction}, {\em (ii) question generation}, and {\em (iii) answer generation}.

Formally, given a sentence $X = x_0, ...,x_n$ with a set of indices $I$ which mark target words (based on  the target extraction heuristics in Section~\ref{targetextraction}),
we aim to produce a set of QA-pairs $(Q_j, A_j)$ using the following pipeline:

1. \textit{Question Prefix Prediction}: 
Let $\Psi$ be the set of all Question Prefixes, each reflecting a relation sense from the list shown in Table~\ref{mapping}.
For each target word \(x_i,\) such that \( i\in I\), we predict a set of possible question prefixes $P_{x_i} \subseteq \Psi$. The set \(P=\bigcup_{i \in I} P_{x_i}\) is now  defined to be the set of all prefixes for all targets in the sentence. 

2.  \textit{Question Generation}: 
For every question prefix \(p\in P\) and all its relevant target words $P_p = \{ x_i |  p \in P_{x_i} \}$, predict question bodies for one or more of the targets $Q_{p}^1, ..., Q_{p}^m $.

3. \textit{Answer Generation}: Let a full question $FQ_p^j$ be defined by the concatenation of the question prefix and the corresponding generated question body  $FQ_p^j = \langle p,Q_{p}^j \rangle $. Given a sentence \(X\) and the  question $FQ_p^j$, we aim to generate an answer \(A_p^j\).

All in all, we can generate up to $|I|\times |\Psi|$ different QAs per sentence.

\subsection{Question Prefix Prediction}
In the first step of our pipeline we are
given a sentence and a marked target, and we aim to predict a set of possible prefixes reflecting potential discourse senses for the relation to be predicted. We frame this task of predicting a set of prefixes as a multi-label classification task.

To represent $I$, the input  sentence $X = x_0, ...,x_n$ is concatenated with a binary target indicator, and special tokens are placed before and after the target $t_i$. 
The output of the system is a set of question prefixes \(P_{x_i}\).

We implement the model using BERT \cite{devlin2019bert} in its standard fine-tuning setting, except that the Softmax layer is replaced by a Sigmoid activation function to support multi-label classification. The predicted question prefixes are obtained by choosing those labels that have a logit $>=\tau=0.3$,
 which was tuned on Dev to maximize UQA F1.
Since the label distribution is skewed, we add weights to the positive examples for the binary cross-entropy loss. 

\subsection{Question Generation}
Next in our pipeline, given the sentence, a question prefix and its relevant targets in the sentence, we aim to generate  question bodies for one or more of the targets. To this end, we employ a Pointer Generator model \cite{jia2016recombination} such that the input to the model is  encoded as follows: $[CLS]$ $x_1, x_2 ... x_n$ $[SEP]$  \(p\) $[SEP]$, with  $p \in P$ being the question prefix. Additionally, we concatenate a target indicator for all relevant targets \(P_p\). The output is one or more question bodies $Q_p$ separated by a delimiter token: $Q_p^1 \: [SEP] \: Q_p^2 \: [SEP] \: ... \: Q_p^m $.

The model then chooses whether to copy a word from the input, or to predict a word during decoding. We use the \textsc{AllenNLP} \cite{Gardner2018AllenNLP} implementation of \textsc{CopyNet} \cite{gu2016incorporating} and adapt it to work with BERT encoding of the input.

\subsection{Answer Generation}
To predict the answer given a full question, we use BERT fine-tuned on SQUAD \cite{rajpurkar2016squad}.\footnote{\url{https://huggingface.co/transformers/pretrained_models.html}} We additionally fine-tune the model on a subset of our training data (all 5004 instances where we could align the answer to a consecutive span in the sentence). Instead of predicting or copying words from the sentence, this  model predicts start and end indices in the sentence.

\section{Results and Discussion}
After running the full pipeline, we evaluate the predicted set of QA-pairs against the gold set using the UQA and LQA metrics, described in section \ref{evalmetrics}. Table~\ref{FullPipeline} shows the results. Note that the LQA is dependent on the UQA, as it calculates the labeled accuracy only for QA pairs that could be aligned with UQA.
The Prefix Accuracy measure complements LQA by evaluating the overall accuracy of predicting a correct question prefix. For this baseline model it shows that generally only half of the generated questions have a question prefix equivalent to gold, leaving room for future models to improve upon. While not comparable, \citet{biran2015pdtb}, for example, mention an F1 of 56.91 for predicting intra-sentential relations.

\begin{table}[t]
\centering
\small
\scalebox{1}{
\begin{tabular}{l|cc}
\hline
& Dev & Test \\  \hline
UQA Precision  & 81.1 & 80.79\\ 
UQA Recall & 84.94 & 86.8\\ 
UQA F1  & 82.98 & 83.69\\ \hline
LQA Accuracy & 67.49 & 66.59\\ 
Prefix Accuracy & 51.3 & 49.94\\ \hline
\end{tabular}}
\caption{Full pipeline performance for the QA-Model evaluated with labeled and unlabeled QA-alignment.}
\label{FullPipeline}
\end{table}
The scores in Table~\ref{AnswerGold} show the results for the subsequent individual steps, given gold input, evaluated using a matching criterion of intersection over union $>=0.5$ with the respective gold span.

\begin{table}[t]
\centering
\small
\scalebox{1}{
\begin{tabular}{l|cc}
\hline
& Dev & Test \\  \hline
Question Prediction & 71.9 &  65.9\\
Answer Prediction &  68.9 &  72.3\\
\hline
\end{tabular}}
\caption{Accuracy of answers predicted by the question and answer prediction model, given a Gold question as input, compared to the Gold spans.}
\label{AnswerGold}
\end{table}

We randomly selected a sample of 50 predicted QAs for a qualitative analysis. 22 instances from this sample were judged as correct, and 2 instances were correct despite not being mentioned in Gold. Examples of good predictions are shown on the left column in Table~\ref{genqex}. 
The model is often able to learn when to do the auxiliary flip from clause to question format and when to change the verb form of the target. Interestingly, whenever the model was not familiar with a specific verb, it chose a similar verb in the correct form, for example `appearing' in Ex.\ 2. The model is also able to form a question using non-adjacent spans of the sentence (Ex.\ 1). Some predictions do not appear in the dataset, but make sense nonetheless. The analysis showed 8 non-grammatical but sensible QAs (i.e. ex.\ 4, where the sense of the relation is still captured), 8 non-sensical but grammatical QAs (ex.\ 5) and 7 QAs that were neither (ex.\ 6). Lastly, we found 3 QAs with good questions and wrong answers. 

\section{Conclusion}
In this work, we show that  discourse relations can be represented as QA pairs. 
This intuitive representation enables scalable, high-quality annotation via crowdsourcing,  which paves the way for learning robust parsers of informational discourse QA pairs. 
In future work, we plan to extend the annotation process to also cover inter-sentential relations.

\section*{Acknowledgments}
We would like to thank Amit Moryossef for his help with the implementation of the frontend, and Julian Michael, Gabriel Stanovsky and the anonymous reviewers for their feedback and suggestions. 
This work was supported in part by grants from Intel Labs, Facebook, the Israel Science Foundation grant 1951/17 and the German Research Foundation through the German-Israeli Project Cooperation (DIP, grant DA 1600/1-1) and by an ERC-StG grant \#677352 and an ISF grant \#1739/26.

\bibliography{anthology,emnlp2020}
\bibliographystyle{acl_natbib}

\appendix

\section{Appendices}
\label{sec:appendix}

\subsection{Reproducibility information}
The code to reproduce the QADiscourse model can be found at \url{https://github.com/ValentinaPy/QADiscourse}.
\paragraph{Calculating the weights for the Loss of the Prefix Classifier}
Each question prefix label is weighted by subtracting the label count from the total count of training instances and dividing it by the label count: $weight_x = total\_instances - count_x / (count_x + 1e-5) $.
\subsection{Annotation Details}
The number of examples and the details of the splits are mentioned in the paper. The data collection process has also been described in the main body of the paper. Here we add a more detailed description of the Target Extraction Algorithm and screenshots of the annotation interfaces.

\paragraph{Target Extraction Algorithm}
In order to extract targets we use the following heuristics:
We split the sentence on the following punctuation: “,” “;” “:”. This provides an initial incomplete segmentation of clauses and subordinate clauses. We will try to find at least one target in each segment.

We then split the resulting text spans from 1. using a set of discourse connectives. We had to remove the most ambiguous connectives from the list, whose tokens might also have other syntactic functions, for example “so, as, to, about”, etc.

We then check the POS tags of the resulting spans and treat each consecutive span of verbs as a target, with the last verb in the consecutive span being the target. In order to not treat cases such as “is also studying” as separate targets, we treat “V ADV V” also as one consecutive span. In case there is no verb in a given span, we chose one of the nouns as the target, but only if the span starts with a discourse connective. This condition allows us to not include nouns as targets that are simply part of enumerations, while at the same time it helps include eventive nouns, see b) for an example. To improve precision (by 0.02) we also excluded the following verbs “said, according, spoke”.

With these heuristics we achieve a Recall of 98.4 and a Precision of 57.4 compared with the discourse relations in Sec. 22 of the PDTB.

\paragraph{Cost Details}
The basic cost for each sentence was 18\textcent, with a bonus of 3\textcent \space for creating a second QA pair and then a bonus of 4\textcent \space for every additional QA pair after the first two. Adjudication was rewarded with 10\textcent \space per sentence. On average 50.3\textcent \space were spent per sentence of Dev and Test, with an average of 2.11 QA pairs per sentence. For Train the average cost per sentence is about 37.1\textcent, with an average of 1.72 QAs.
\newpage
\paragraph{Annotation Interfaces}
The following screenshots display the Data Collection and Adjudication interfaces.

\begin{figure}[!htb]
\includegraphics[width=3in, left]{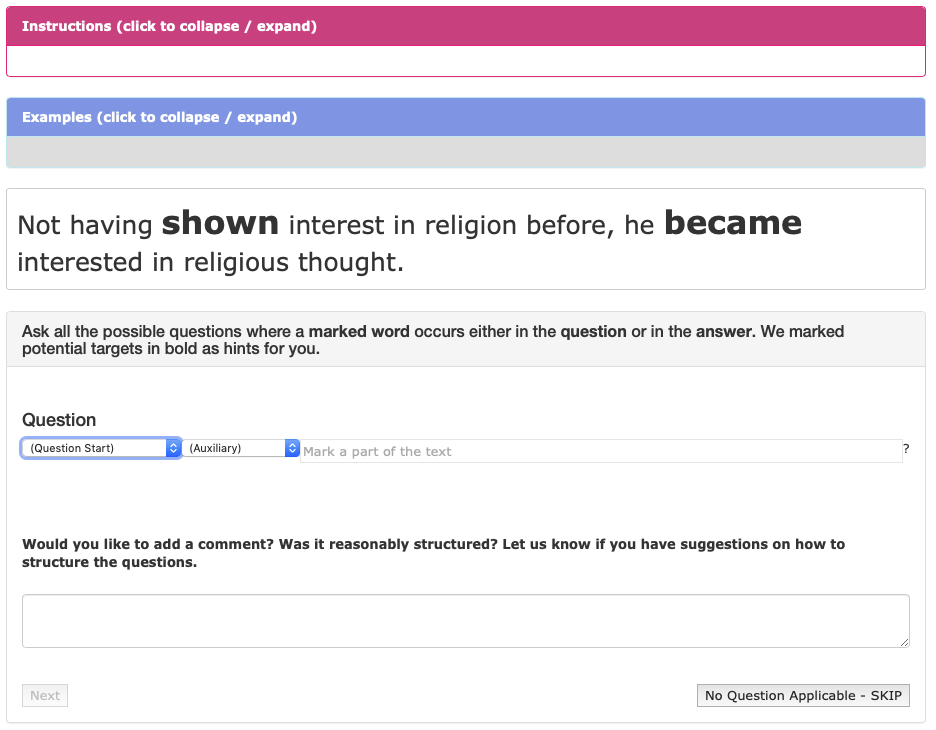}
\caption{Interface for the Question Generation step.}
\label{qgen}
\end{figure}

\begin{figure}[!htb]
\includegraphics[width=3in, left]{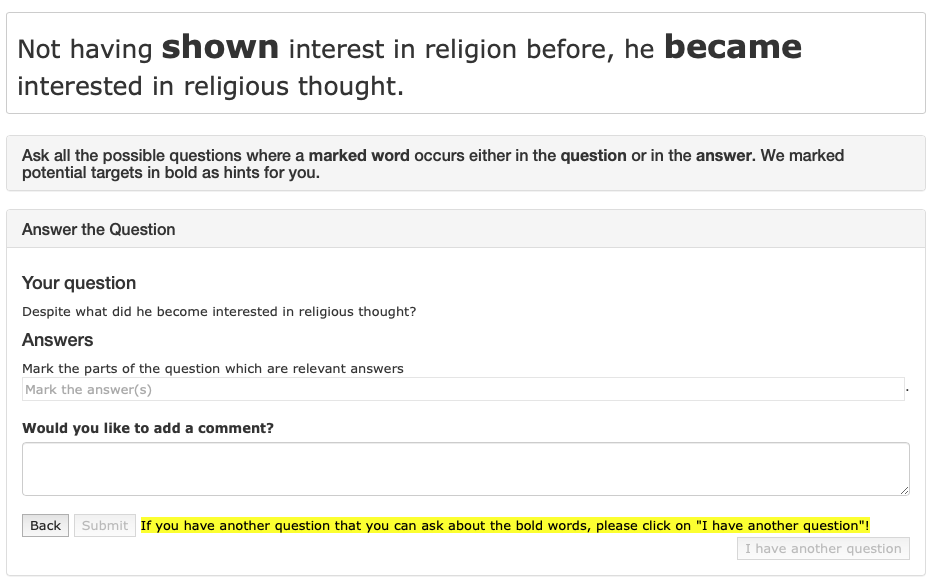}
\caption{Interface for the Answer Generation step.}
\label{agen}
\end{figure}

\begin{figure}[!htb]
\includegraphics[width=3in, left]{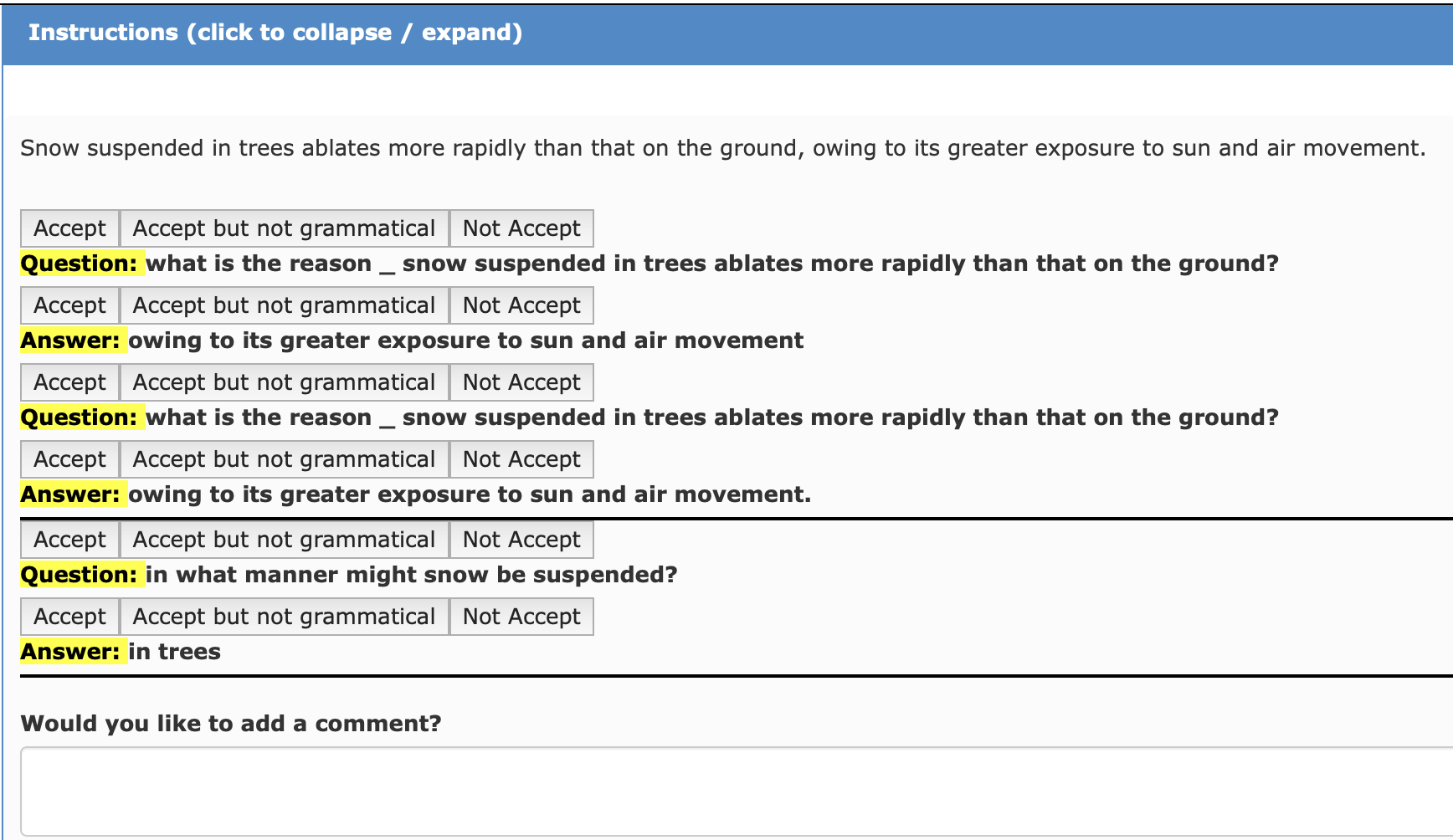}
\caption{Interface for the Adjudication step.}
\label{consolidation}
\end{figure}

\subsection{Data Examples}
\begin{table*}[htb!]
\resizebox{\textwidth}{!}{%
    \centering
    \small
    \begin{tabular}{|p{2.85in}|p{2in}|p{1in}|} \hline
    
           \textbf{Sentence} & \textbf{Question} & \textbf{Answer}\\ \hline
           
       An inquest found he'd committed suicide, but some dispute this and believe it was an accident . & Instead of what do some believe it was an accident? & suicide\\ \hline
       
              On Sunday, in a video posted on YouTube, Anonymous announced their intentions saying, ``From the time you have received this message, our attack protocol has past been executed and your downfall is underway." & Since when has our attack protocol past been executed and your downfall is underway? & From the time you have received this message\\ \hline
              
      It is unclear why this diet works. & Despite what does this diet work? & Being unclear why \\ \hline  
      
            It's a downgraded budget from a downgraded Chancellor [...] Debt is higher in every year of this Parliament than he forecast at the last Budget. & What is similar to it's a downgraded budget? & It's a downgraded Chancellor \\ \hline  
            
                       According to Pakistani Rangers, the firing from India was unprovoked in both Sunday and Wednesday incidents; Punjab Rangers in the first incident, and Chenab Rangers in the second incident, retaliated with intention to stop the firing. & What is the reason punjab Rangers and Chenab Rangers retaliated? & with intention to stop the firing \\ \hline  
                       
           The vessel split in two and is leaking fuel oil .& After what did the vessel leak fuel oil? & The vessel split in two \\ \hline  
           
           In contrast to the predictions of the Met Office, the Environment Agency have said that floods could remain in some areas of England until March, and that up to 3,000 homes in the Thames Valley could be flooded over the weekend.& What is contrasted with the predictions of the Met Office? & the Environment Agency have said that floods could remain in some areas of England until March, and that up to 3,000 homes in the Thames Valley could be flooded over the weekend \\ \hline  
    \end{tabular}}
    \caption{Examples for QA pairs that were annotated in the dataset.}
    \label{corpusex}
\end{table*}

\begin{table*}[htb!]
\resizebox{\textwidth}{!}{%
    \centering
    \small
    \begin{tabular}{|p{2.85in}|p{2in}|p{1in}|} \hline
    
           \textbf{Sentence} & \textbf{Question} & \textbf{Answer}\\ \hline
           
       Standard addition can be applied to most analytical techniques and is used instead of a calibration curve to solve the matrix effect problem. & Instead of what is standard addition used? & a calibration curve\\ \hline
       
              State officials therefore share the same interests as owners of capital and are linked to them through a wide array of social, economic, and political ties. & What is similar to state officials share the same interests as owners of capital? & are linked to them through a wide array of social, economic, and political ties\\ \hline
              
      Recently, this field is rapidly progressing because of the rapid development of the computer and camera industries. & What is the reason this field is rapidly progressing? & Because of the rapid development of the computer and camera industries \\ \hline  
      
            Civilization was the product of the Agricultural Neolithic Revolution; as H. G. Wells put it, ``civilization was the agricultural surplus.'' & In what manner was civilization the product of the Agricultural Neolithic Revolution? & civilization was the agricultural surplus \\ \hline  
            
                       The portrait shows such ruthlessness in Innocent's expression that some in the Vatican feared that Velázquez would meet with the Pope's displeasure, but Innocent was well pleased with the work, hanging it in his official visitor's waiting room. & Despite what was Innocent well pleased with The work? & The portrait shows such ruthlessness in Innocent's expression that some in the Vatican feared that Velázquez would meet with the Pope's displeasure \\ \hline  
                       
           All tropical cyclones lose strength once they make landfall.& After what do tropical cyclones lose strength? & once they make landfall \\ \hline  
           
           The investigation, led by former Dutch General Patrick Cammaert , is separate from the investigation led by the UN's Human Rights Council.& What is contrasted with the investigation, led by former Dutch General Patrick Cammaert? & the investigation led by the UN's Human Rights Council \\ \hline  
    \end{tabular}}
    \caption{Examples for QA pairs that were predicted with the full pipeline.}
    \label{pdtbexamples}
\end{table*}

\begin{table*}[htb!]
\resizebox{\textwidth}{!}{%
    \centering
    \small
    \begin{tabular}{|p{2in}|p{1in}|p{1in}|p{2in}|} \hline
    
           \textbf{Sentence} & \textbf{Question} & \textbf{Answer} & \textbf{PDTB senses}\\ \hline
           
       It was ``the Soviets' Vietnam." The Kabul regime would fall. & After what would the Kabul regime fall? & after the Soviets' Vietnam & Expansion.Conjunction\\ \hline
       
              Eight months after Gen. Boris Gromov walked across the bridge into the U.S.S.R., a Soviet-controlled regime remains in Kabul, the refugees sit in their camps, and the restoration of Afghan freedom seems as far off as ever. & What is the reason the restoration of Afghan freedom seems as far off as ever? & a Soviet-controlled regime remains in Kabul & Temporal.Asynchronous.Succession, Expansion.Conjunction\\ \hline
              
      Soviet leaders said they would support their Kabul clients by all means necessary--and did. & In what manner would soviet leaders support their Kabul clients? & soviet leaders said they would support their kabul clients by all means necessary &  Expansion.Conjunction\\ \hline  
      
      Soviet leaders said they would support their Kabul clients by all means necessary--and did. & After what did Soviet leaders support their Kabul clients by all means necessary? & after soviet leaders said they would & Expansion.Conjunction \\ \hline  
      
      With the February 1987 U.N. accords ``relating to Afghanistan," the Soviet Union got everything it needed to consolidate permanent control. & In what manner did the Soviet Union get everything it needed to consolidate permanent control? & with the February 1987 u.n. accords  ``relating to Afghanistan,"& Contingency.Cause.Reason, Contingency.Purpose.Arg2-as-goal \\ \hline  
            
      The terms of the Geneva accords leave Moscow free to provide its clients in Kabul with assistance of any kind--including the return of Soviet ground forces--while requiring the U.S. and Pakistan to cut off aid. & What is the result of the terms of the Geneva accords? & leaving Moscow free to provide its clients in Kabul with assistance of any kind while requiring the U.S. and Pakistan to cut off aid& Temporal.Synchronous, Comparison.Contrast\\ \hline  
                       
      The only fly in the Soviet ointment was the last-minute addition of a unilateral American caveat, that U.S. aid to the resistance would continue as long as Soviet aid to Kabul did.& What is the reason for the only fly in the Soviet ointment? & the last-minute addition of a unilateral American caveat, that U.S. aid to the resistance would continue as long as Soviet aid to Kabul did & Expansion.Level-of-detail.Arg2-as-detail, Contingency.Condition.Arg2-as-cond, Temporal.Synchronous \\ \hline  
           
      But as soon as the accords were signed, American officials sharply reduced aid.& In what manner did American officials reduce aid? & American officials sharply reduced aid & Temporal.Asynchronous.Succession \\ \hline  
      
      Moscow claims this is all needed to protect the Kabul regime against the guerrilla resistance.& What is the reason Moscow claims this is all needed?& to protect the Kabul regime against the guerrilla resistance & Contingency.Condition.Arg2-as-cond \\ \hline  
            
      But this is not the entire Afghan army, and it is no longer Kabul's only military force.&What is similar to it not being the entire Afghan army? & is no longer Kabul's only military force. & Expansion.Conjunction \\ \hline  
                  
      The deal fell through, and Kandahar remains a major regime base.& After what did Kandahar remain a major regime base? & after the deal fell through & Contingency.Cause.Result, Expansion.Conjunction \\ \hline  
      
            The deal fell through, and Kandahar remains a major regime base.& Since when does Kandahar remain a major regime base? & since the deal fell through & Contingency.Cause.Result, Expansion.Conjunction \\ \hline 
      
      The wonder is not that the resistance has failed to topple the Kabul regime, but that it continues to exist and fight at all.& Despite what is the wonder that it continues to exist and fight at all? & despite the resistance failing to topple the kabul regime & Comparison.Contrast, Expansion.Substitution.Arg2-as-subst \\ \hline  
      
      Last summer, in response to congressional criticism, the State Department and the CIA said they had resumed military aid to the resistance months after it was cut off; but it is not clear how much is being sent or when it will arrive.& what is the result of congressional criticism last summer? & the state department and the CIA said they had resumed military aid to the resistance & Temporal.Asynchronous.Succession, Comparison.Concession.Arg2-as-denier, Expansion.Conjunction \\ \hline  
    \end{tabular}}
    \caption{Examples for additional relations expressed through QA pairs that do not appear in the PDTB, Part 1.}
    \label{pdtb_anno0}
\end{table*}

\begin{table*}[htb!]
\resizebox{\textwidth}{!}{%
    \centering
    \small
    \begin{tabular}{|p{2in}|p{1in}|p{1in}|p{2in}|} \hline
    
           \textbf{Sentence} & \textbf{Question} & \textbf{Answer} & \textbf{PDTB senses}\\ \hline
           
                 Beyond removing a competitor, the combination should provide `` synergies," said Fred Harlow, Unilab's chief financial officer.& While what should the combination provide synergies? & removing a competitor. & Expansion.Conjunction \\ \hline  

      In Los Angeles, for example, Central has had a strong market position while Unilab's presence has been less prominent, according to Mr. Harlow.& In what case has Central had a strong market position while Unilab's presence has been less prominent? & in Los Angeles & Comparison.Contrast, Temporal.Synchronous \\ \hline 
           
      A Daikin executive in charge of exports when the high-purity halogenated hydrocarbon was sold to the Soviets in 1986 received a suspended 10-month jail sentence.& What is the result of the high-purity halogenated hydrocarbon being sold to the Soviets in 1986? & a Daikin executive in charge of exports received a suspended 10-month jail sentence & Temporal.Synchronous \\ \hline 
      
      In Los Angeles, for example, Central has had a strong market position while Unilab's presence has been less prominent, according to Mr. Harlow.& in what case has central had a strong market position while Unilab's presence has been less prominent? & in Los Angeles & Comparison.Contrast, Temporal.Synchronous \\ \hline 
      
      Mr. Mehl noted that actual rates are almost identical on small and large-denomination CDs, but yields on CDs aimed at the individual investor are boosted by more frequent compounding.& In what manner are yields on CDs aimed at the individual investor boosted?& by more frequent compounding& Comparison.Concession.Arg2-as-denier\\ \hline 

      Judge Masaaki Yoneyama told the Osaka District Court Daikin's ``responsibility is heavy because illegal exports lowered international trust in Japan." Sale of the solution in concentrated form to Communist countries is prohibited by Japanese law and by international agreement.& Except when is the solution in concentrated form sold? & except to communist countries & Contingency.Cause.Reason, EntRel\\ \hline 
      
      Japan has supported a larger role for the IMF in developing-country debt issues, and is an important financial resource for IMF-guided programs in developing countries.& In what case is Japan an important financial resource for imf-guided programs? & in developing countries & Expansion.Conjunction\\ \hline 
      
      Japan has supported a larger role for the IMF in developing-country debt issues, and is an important financial resource for IMF-guided programs in developing countries.& While what has Japan supported a larger role for the IMF in developing-country debt issues? & while it is an important financial resource for imf-guided programs in developing countries & Expansion.Conjunction\\ \hline 
      
      The last U.S. congressional authorization, in 1983, was a political donnybrook and carried a \$6 billion housing program along with it to secure adequate votes.& What is an example of something being a political donnybrook? & the last u.s. congressional authorization, in 1983 & Contingency.Purpose.Arg2-as-goal, Expansion.Conjunction\\\hline 
      
      Instead, the tests will focus heavily on new blends of gasoline, which are still undeveloped but which the petroleum industry has been touting as a solution for automobile pollution that is choking urban areas.& What is the reason tests will focus heavily on new blends of gasoline? & the petroleum industry has been touting as a solution for automobile pollution & Comparison.Concession.Arg2-as-denier\\\hline 
      
      While major oil companies have been experimenting with cleaner-burning gasoline blends for years, only Atlantic Richfield Co. is now marketing a lower-emission gasoline for older cars currently running on leaded fuel.& While what is Atlantic Richfield co. marketing a lower-emission gasoline for older cars currently running on leaded fuel? & while major oil companies have been experimenting with cleaner-burning gasoline blends & Comparison.Contrast\\\hline 
    \end{tabular}}
    \caption{Examples for additional relations expressed through QA pairs that do not appear in the PDTB, Part 2.}
    \label{pdtb_anno_1}
\end{table*}

\begin{table*}[htb!]
\resizebox{\textwidth}{!}{%
    \centering
    \small
    \begin{tabular}{|p{2in}|p{1in}|p{1in}|p{2in}|} \hline
    
           \textbf{Sentence} & \textbf{Question} & \textbf{Answer} & \textbf{PDTB senses}\\ \hline
            Instead, a House subcommittee adopted a clean-fuels program that specifically mentions reformulated gasoline as an alternative.& What is the result of a house subcommittee adopting a clean-fuels program?& reformulated gasoline as an alternative.& Expansion.Level-of-detail.Arg2-as-detail\\\hline 
      
      The Bush administration has said it will try to resurrect its plan when the House Energy and Commerce Committee takes up a comprehensive clean-air bill.&In what case will the Bush administration try to resurrect its plan?& when the house energy and commerce committee takes up a comprehensive clean-air bill& Temporal.Synchronous\\\hline 
      
      That compares with per-share earnings from continuing operations of 69 cents the year earlier; including discontinued operations, per-share was 88 cents a year ago.& In what manner does that compare with per-share earnings from continuing operations of 69 cents the year earlier?& including discontinued operations, per-share was 88 cents a year ago.& Comparison.Contrast\\\hline 
      
      Analysts estimate Colgate's sales of household products in the U.S. were flat for the quarter, and they estimated operating margins at only 1\% to 3\%&While what did analysts estimate Colgate's sales of household products in the U.S. were flat for the quarter?& they estimated operating margins at only 1\% to 3\%& Expansion.Conjunction\\\hline 
      
      Analysts estimate Colgate's sales of household products in the U.S. were flat for the quarter, and they estimated operating margins at only 1\% to 3\%&After what did analysts estimate Colgate's sales of household products in the U.S. were flat?& after the quarter& Expansion.Conjunction\\\hline 
      
      The programs will be written and produced by CNBC, with background and research provided by staff from U.S. News & What is similar to the programs being written by CNBC?& being produced by CNBC& Expansion.Conjunction\\\hline 
      
      The programs will be written and produced by CNBC, with background and research provided by staff from U.S. News & In what manner will background and research be provided for the programs?& by staff from U.S. news & Expansion.Conjunction\\\hline 
      
      The programs will be written and produced by CNBC, with background and research provided by staff from U.S. News & In what manner will the programs be written and produced?& the programs will be written and produced by CNBC, with background and research provided by staff from U.S. news& Expansion.Conjunction\\\hline 
      
      Frank Carlucci III was named to this telecommunications company's board, filling the vacancy created by the death of William Sobey last May.& After what was Frank Carlucci III named to this telecommunications company\'s board?& the death of William Sobey last may& Contingency.Cause.Result\\\hline 
      
      Weyerhaeuser's pulp and paper operations were up for the nine months, but full-year performance depends on the balance of operating and maintenance costs, plus pricing of certain products, the company said.& What is contrasted with full-year performance of Weyerhaeuser's pulp and paper operations?& nine month performance& Comparison.Concession.Arg2-as-denier\\\hline 
      
      Weyerhaeuser's pulp and paper operations were up for the nine months, but full-year performance depends on the balance of operating and maintenance costs, plus pricing of certain products, the company said.& What is the result of Weyerhaeuser's full-year performance?& depends on the balance of operating and maintenance costs, plus pricing of certain products.& Comparison.Concession.Arg2-as-denier\\\hline 
    \end{tabular}}
    \caption{Examples for additional relations expressed through QA pairs that do not appear in the PDTB, Part 3.}
    \label{pdtb_annos_ex}
\end{table*}

\end{document}